\documentclass{article}

\usepackage{algorithm}
\usepackage{algorithmic}
\usepackage{microtype}
\usepackage{graphicx}
\usepackage{caption}
\usepackage{subcaption}
\usepackage{booktabs} 
\usepackage{amsmath}
\usepackage{amssymb}
\usepackage{natbib}

\usepackage{hyperref}

\usepackage[accepted]{icml2019}

\icmltitlerunning{Intrinsically Sparse Long Short-Term Memory Networks}

\begin{document}

\twocolumn[
\icmltitle{Intrinsically Sparse Long Short-Term Memory Networks}



\icmlsetsymbol{equal}{*}

\begin{icmlauthorlist}
\icmlauthor{Shiwei Liu}{to}
\icmlauthor{Decebal Constantin Mocanu}{to}
\icmlauthor{Mykola Pechenizkiy}{to}
\end{icmlauthorlist}
\icmlaffiliation{to}{Department of Mathematics and Computer Science, Eindhoven University of Technology, Netherlands}

\icmlcorrespondingauthor{Shiwei Liu}{s.liu3@tue.nl}

\icmlkeywords{scalable deep learning, adaptive sparse connectivity, sparse evolutionary training,  curse of dimensionality, microarray gene expression, high-dimensional data }

\vskip 0.3in
]



\printAffiliationsAndNotice{}  

\begin{abstract}
Long Short-Term Memory (LSTM) has achieved state-of-the-art performances on a wide range of tasks. Its outstanding performance is guaranteed by the long-term memory ability which matches the sequential data perfectly and the gating structure controlling the information flow. However, LSTMs are prone to be memory-bandwidth limited in realistic applications and need an unbearable period of training and inference time as the model size is ever-increasing. To tackle this problem, various efficient model compression methods have been proposed. Most of them need a big and expensive pre-trained model which is a nightmare for resource-limited devices where the memory budget is strictly limited. To remedy this situation, in this paper, we incorporate the Sparse Evolutionary Training (SET) procedure into LSTM, proposing a novel model dubbed SET-LSTM. Rather than starting with a fully-connected architecture, SET-LSTM has a sparse topology and dramatically fewer parameters in both phases, training and inference. Considering the specific architecture of LSTMs, we replace the LSTM cells and embedding layers with sparse structures and further on, use an evolutionary strategy to adapt the sparse connectivity to the data. Additionally, we find that SET-LSTM can provide many different good combinations of sparse connectivity to substitute the overparameterized optimization problem of dense neural networks. Evaluated on four sentiment analysis classification datasets, the results demonstrate that our proposed model is able to achieve usually better performance than its fully connected counterpart while having less than 4\% of its parameters. 
\end{abstract}

\section{Introduction}
\label{submission}

In recent years, Long Short-Term Memory (LSTM) has returned to people's attention with its outstanding performance in speech recognition \cite{graves2013hybrid}, neural machine translation \cite{sutskever2014sequence}, sentiment classification \cite{yang2016hierarchical} and other tasks related to sequential data. LSTM's success is due to its two-fold surprising properties. The first one is the intrinsic ability to memorize historical information, which fits very well with sequential data. This ability is its main advantage compared with other mainstream networks such as Multilayer Perceptron (MLP), Convolutional Neural Network (CNN). Second, the exploding and vanishing gradient problems are eased through memory gates controlling the flow of information according to the different objectives. Moreover, mixed models obtained by stacking LSTM layers together with other type of neural networks layers can improve state-of-the-art in various applications.

However, the large LSTM-based models are often associated with expensive computations, large memory requests and inefficient processing time in both phases, training and inference. For example, around 30\% of the Tensor Processing Unit (TPU) workload in the Google cloud is caused by LSTMs \cite{jouppi2017datacenter}. The computation-intensive and memory-intensive are at odds with the trend of deploying these powerful models on resource-limited devices. Different from other neural networks, LSTMs are relatively more challenging to be compressed due to the complicated architecture that the information gained from one cell will be shared across all the time steps \cite{wen2017learning}. Despite this challenge, researchers already proposed many effective methods to address this problem, including Sparse Variational Dropout (Sparse VD) \cite{lobacheva2017bayesian}, sparse regularization \cite{wen2017learning}, distillation \cite{tian2017deep}, low-rank factorizations and parameter sharing \cite{lu2016learning} and pruning \cite{han2017ese, narang2017exploring, lee2018snip}, etc. All of them can achieve promising compression rates with negligible performance loss. Nonetheless, one common shortcoming hindering their applications on resource-limited device is that expensive fully-connected networks are needed at the beginning. Such very large pre-trained models where most layers are fully-connected (FC) are prone to be memory bound in realistic applications \cite{jouppi2017datacenter}. At the same time, \cite{mocanu2018scalable} have proposed the Sparse Evolutionary Training (SET) procedure, which creates sparsely connected layers before training. Such layers start from an Erd\H{o}s-R\'{e}nyi random graph connectivity, and use an evolutionary training strategy to force the sparse connectivity to fit the data during the training phase.

In this paper, we introduce adaptive sparse connectivity into the LSTM world. Concretely, we propose a new  sparse LSTM model trained with SET, and dubbed further SET-LSTM. In comparison with all LSTM variants discussed above, SET-LSTM is sparse from the design phase, before training. Considering the specific structure inside LSTM cells, we first replace the fully-connected layers within the LSTM cells. Secondly, we sparsify the embedding layer to further reduce a major number of parameters as it is usually the largest layer in LSTMs. Evaluated on four sentiment classification datasets, our proposed model is able to achieve higher accuracy than fully-connected LSTMs on three of them and just a bit lower accuracy on the last one, while having about 25 times less parameters. To understand the beneficial effect of adaptive sparse connectivity on model performance, we study the sparsely connected layers topologies obtained after the training process, and we show that even if in terms of accuracy the results are similar, the topologies are completely different. This suggests that adaptive sparse connectivity may be a way to avoid the overparameterized optimization problem of fully-connected neural networks, as it yields many amenable local optima.

\section{Preliminaries}
\subsection{LSTM Compression}
There are various effective techniques to shrink the size of large LSTMs, at the same time, preserving the competitive performance. Here, we divide them into pruning methods and non-pruning methods. 

\textbf{Pruning methods.}
Pruning as a classical model compression method has been widely used to different models successfully such as MLPs, CNNs and LSTMs. By eliminating the unimportant weights based on a certain criterion, pruning is able to achieve high compression ratio without substantial loss in accuracy. Pruning-based LSTMs compression methods can be categorized into two branches: post-training and direct sparse training, according to whether an expensive fully-connected network is needed before the training process.

Pruning from a fully-connected network is an overwhelming branch to compress neural networks. \cite{giles1994pruning} proposes a simple pruning and retraining strategy to recurrent neural networks (RNNs). However, the inevitably expensive computation and prohibitively many training iterations are the main disadvantages of these methods. Recently, \cite{han2015learning} makes pruning stand out from other methods by pruning the magnitude of weights and retraining the network. Based on the pruning approach of \cite{han2015learning}, \cite{han2017ese} proposes an efficient method to compress LSTMs by combining pruning with quantization together. On the other hand, \cite{narang2017exploring} shrinks the post-pruning sparse LTSM size by 90\% through a monotonically increasing threshold. Using a set of hyperparameters is able to determine the specific thresholds for different layers. \cite{lobacheva2017bayesian} applies Sparse VD to LSTM and achieves 99.5\% sparsity from the perspective of Bayesian networks. Despite the success of post-training, an expensive fully-connected network is required at the beginning stage, which leads to inevitable memory requirement and computation cost. 

As an emerging branch, direct sparse training can effectively avoid the dependence on the original large networks. Nest \cite{dai2017nest} gets rid of an original huge network by a grow-and-prune paradigm, that is, expanding a small randomly initialized network to a large one and then shrink it down. However, it will not be feasible under a really strict parameters budget. \cite{bellec2017deep} proposes deep rewiring (DEEP R) that guarantees the strictly limited connections by adding a hard constraint to a sample process based on which the sparse connection is rewired. Different from sampling network architecture, our approach use an evolutionary way to dynamically change the topology based on the importance of connections. \cite{mostafa2019parameter} proposes a direct sparse training technique via dynamic sparse reparameterization. Heuristically, it uses a global threshold to prune the magnitude of weights. 

\textbf{Non-pruning methods.}
In addition to pruning, other approaches also make significant contribution to LSTMs compression, including distillation \cite{tian2017deep}, matrix factorization \cite{kuchaiev2017factorization}, parameter sharing \cite{lu2016learning}, group Lasso regularization \cite{wen2017learning}, weight quantization \cite{zen2016fast}, etc.
\subsection{Sparse Evolutionary Training}
Sparse Evolutionary Training (SET) \cite{mocanu2018scalable} is a simple but efficient algorithm which is able to train a directly sparse neural network with no decrease of accuracy. SET algorithm is given in Algorithm \ref{SET}. It does not start from a large fully-connected network. Instead, the random initialization by an Erd{\H{o}}s-R{\'e}nyi topology makes it possible to handle situations where the parameters budget is extremely limited from beginning to end. And given that the random initialization may not be suitable for the data distribution, a fraction $\zeta$ of the connections with the smallest weights will be pruned and an equal number of novel connections will be grown after each epoch. This evolutionary training is capable of guaranteeing a constant sparsity level during the whole learning process and to help in preventing overfitting. The connection (${W}^k_{ij}$) between neuron $h_j^{k-1}$ and $h_i^k$ exists with the probability:  
\begin{equation}
 p({W}^k_{ij})=\frac{\epsilon(n^k+n^{k-1})}{n^kn^{k-1}}
 \label{Eq:probtopology}
\end{equation}
where $n^{k}, n^{k-1}$ are the number of neurons of layer $h^{k}$ and $h^{k-1}$, respectively; $\epsilon$ is a parameter determining the sparsity level. Apparently, the smaller $\epsilon$ is, the more sparse the network is. The connections between the two layers are collected in a sparse weight
matrix $\textbf{W}^k \in \textbf{R}^{n^{k-1}\times{n^{k}}}$. Compared with fully-connected layers whose number of connections is $n^{k}n^{k-1}$ , the SET sparse layers only have $n^W = \mid \textbf{W}^k \mid = \epsilon(n^{k}+n^{k-1})$ connections which can significantly alleviate the pressure of the expensive memory footprint. It is worth noting that, during the learning phase, the initial topology would evolve toward to a scale-free one. 
\makeatletter
\def\BState{\State\hskip-\ALG@thistlm}
\makeatother
\begin{algorithm}
\caption{SET pseudocode}\label{SET}
\begin{algorithmic}[1]
\STATE \%\textit{Sparse Topology Initializaiton};
\STATE initialize ANN model;
\STATE set $\epsilon$ and $\zeta$; 
\FOR{\textit{each bipartite fully-connected layer of the ANN}} 
    \STATE replace FC layer with Sparse Connected(SC) layer with a Erd\H{o}s-R\'{e}nyi topology given by $\epsilon$ and Eq.\ref{Eq:probtopology};
\ENDFOR 
\STATE initialize training algorithm parameters;
\STATE \%\textit{Training};
\FOR{\textit{each training epoch \textit{i}}}
\STATE perform standard training procedure;
\STATE perform weights update;
\FOR{\textit{each bipartite SC layer of the ANN}}
    \STATE remove a fraction $\zeta$ of the smallest positive weights;
    \STATE remove a fraction $\zeta$ of the largest negative weights;
    \IF{\textit{\textit{i} is not the last training epoch}}
      \STATE add randomly new weights (connections) in the same amount as the ones removed previously;
    \ENDIF
\ENDFOR
\ENDFOR
\vskip -0.2in
\end{algorithmic}
\end{algorithm}

\section{SET-LSTM}
In this section, we describe our proposed SET-LSTM model, and how we apply SET to compress the LSTM cells and the embedding layer.
\subsection{SET-LSTM Cells}
\begin{figure}[ht]
\vskip 0 in
\begin{center}-eps-converted-to
\centerline{\includegraphics[width=\columnwidth]{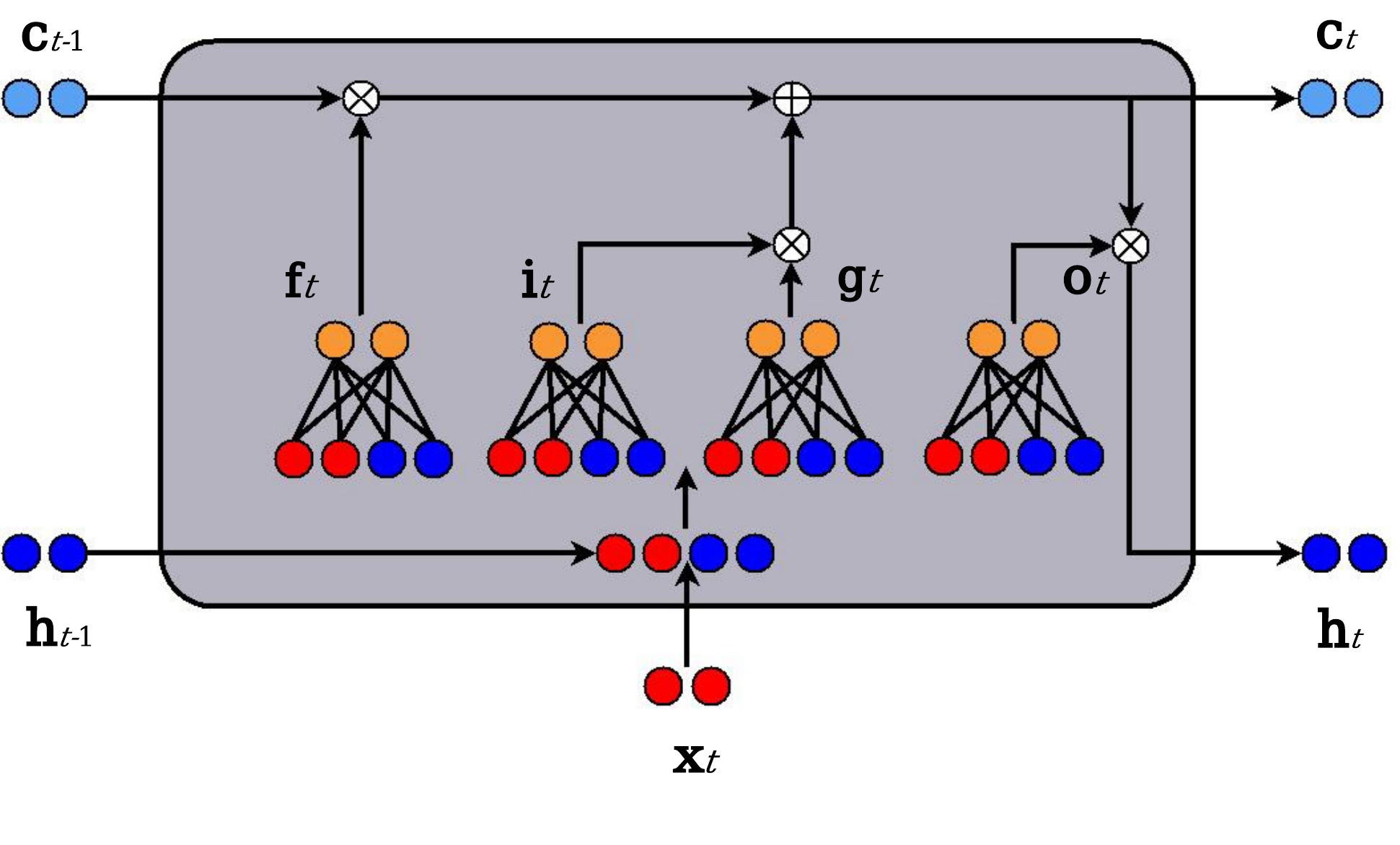}}
\vskip -0.2in
\caption{Schematic diagram of the LSTM cell}
\label{Fig:SLTMCELL}
\end{center}
\vskip 0in
\end{figure}

\begin{figure}[ht]
\vskip -0.2in
\begin{center}
\centerline{\includegraphics[width=\columnwidth]{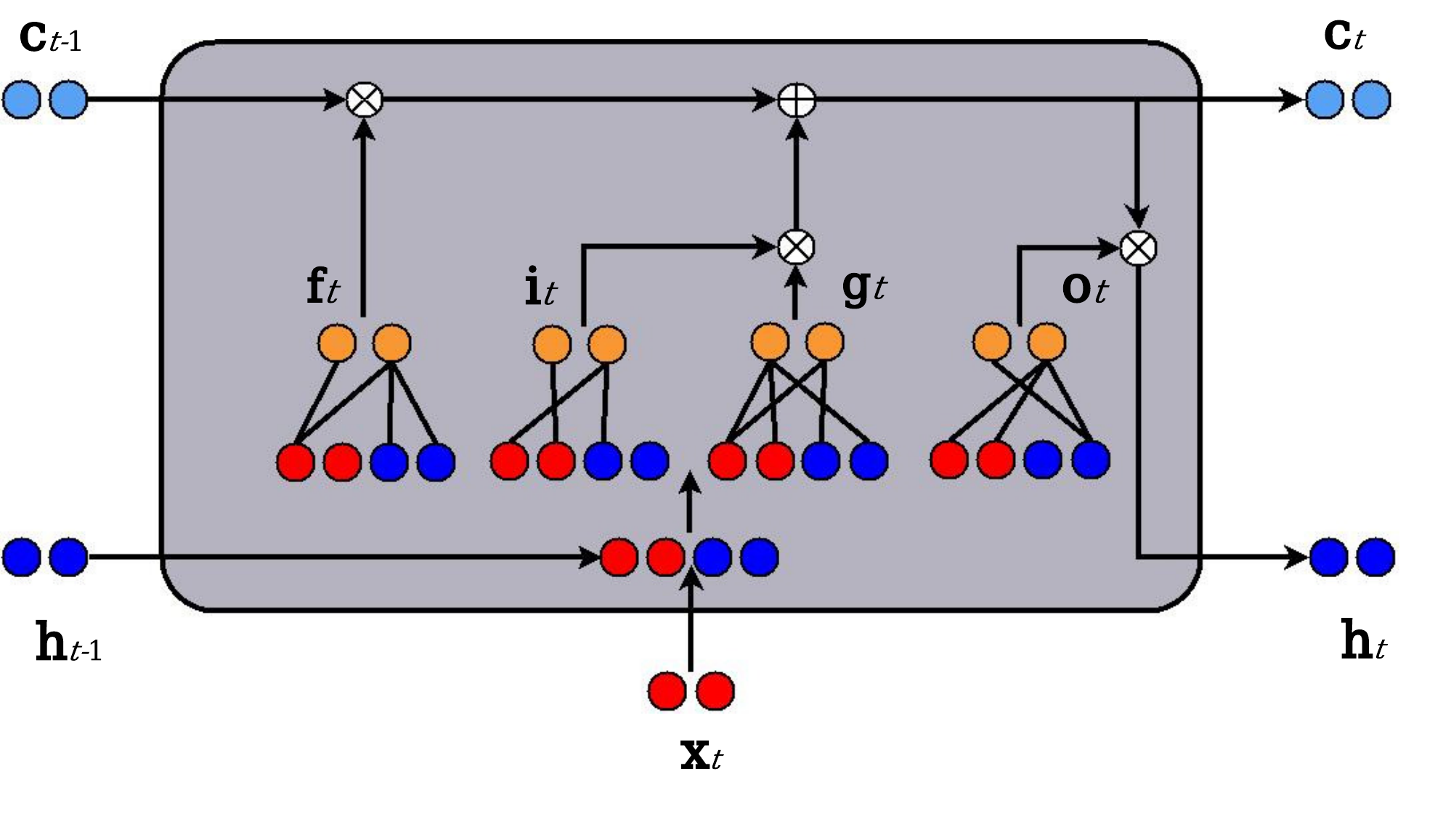}}
\vskip -0.2in
\caption{Schematic diagram of the SET-LSTM cell}
\label{Fig:SparseLstmCell}
\end{center}
\vskip -0.2in
\end{figure}

\begin{figure*}[ht]
\vskip 0.1in
\begin{center}
\centerline{\includegraphics[width=140mm,height=70mm]{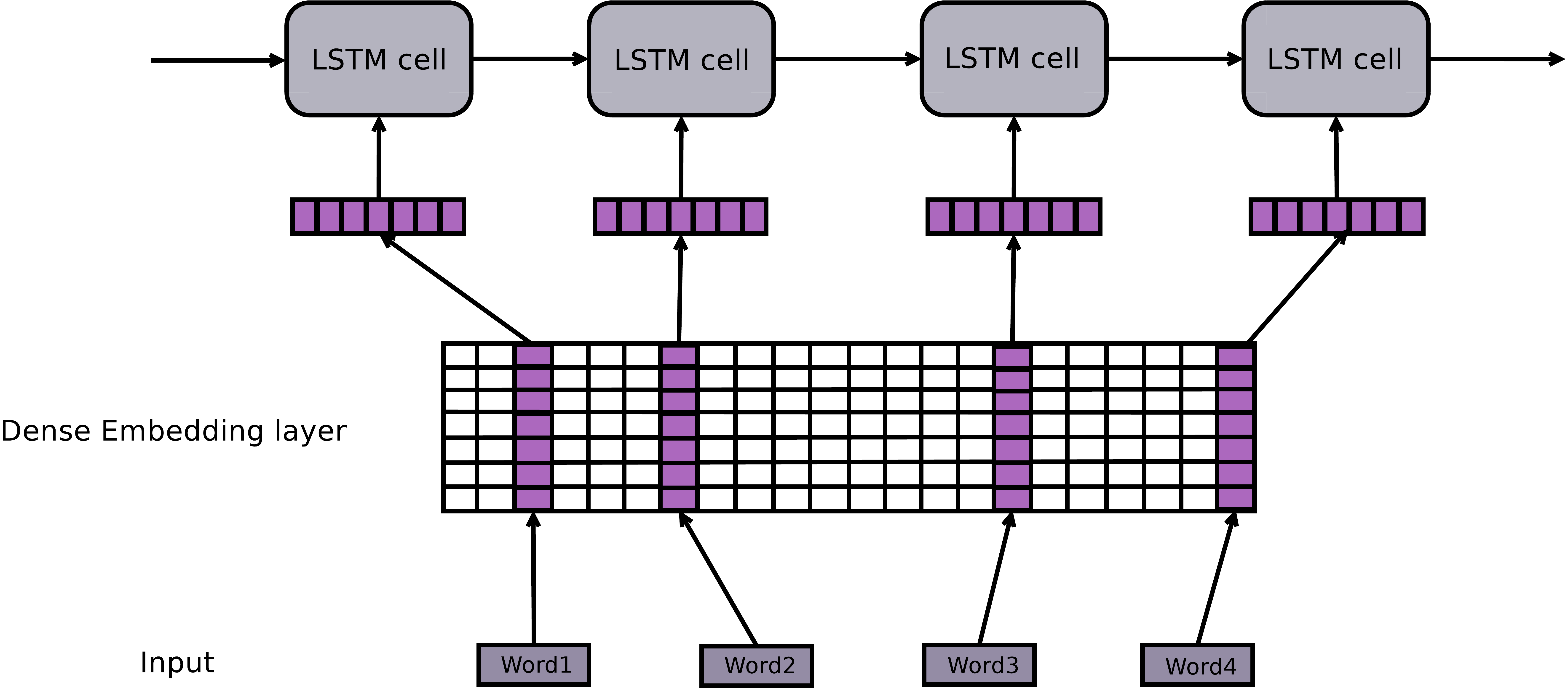}}
\caption{Dense embedding layer}
\label{Fig:dense_embedding}
\end{center}
\vskip -0.2in
\end{figure*}
  
\begin{figure*}[ht]
\vskip 0.1in
\begin{center}
\centerline{\includegraphics[width=140mm,height=70mm]{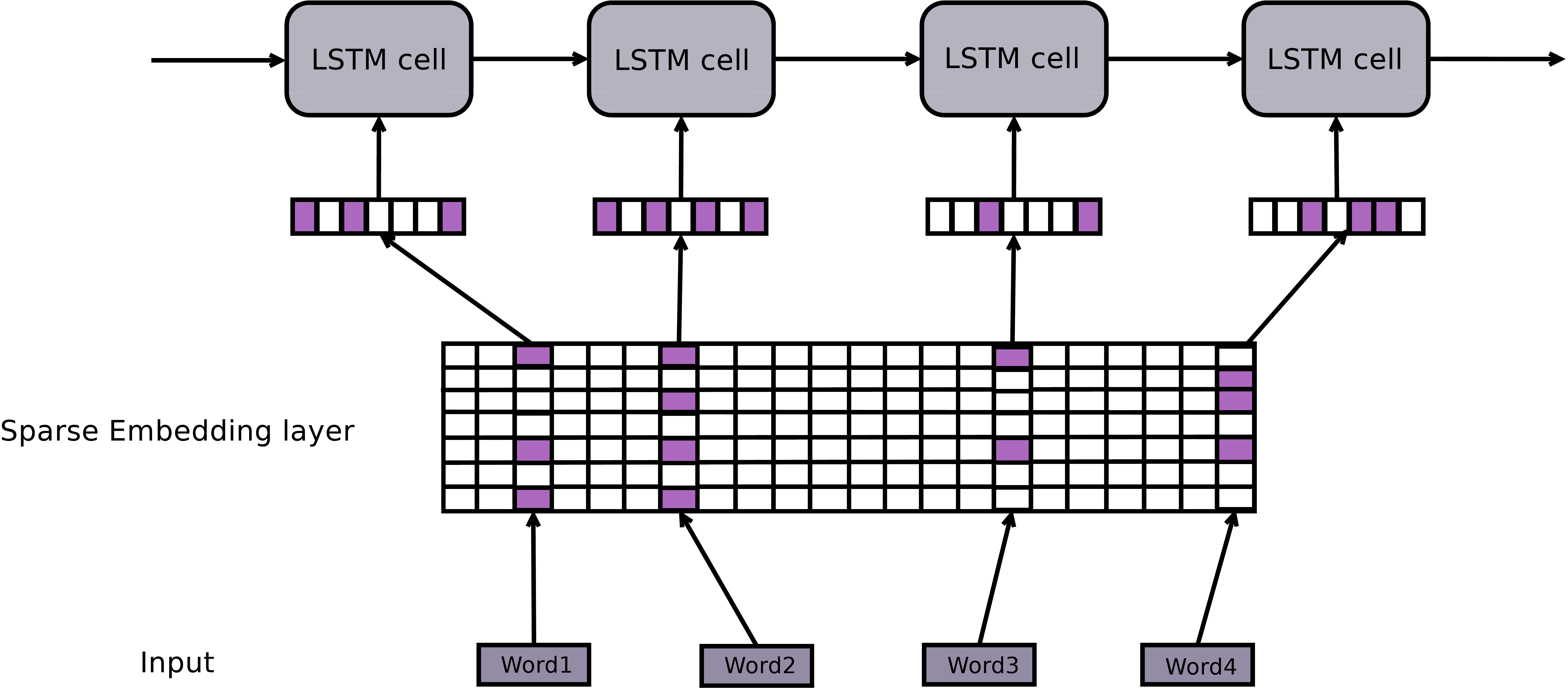}}
\caption{SET sparse embedding layer}
\label{Fig:sparse_embedding}
\end{center}
\vskip -0.2in
\end{figure*}

The conventional schematic of the LSTM cell is shown in Figure~\ref{Fig:SLTMCELL}. The gates ($f_{t}$, $i_{t}$, $g_{t}$ and $o_{t}$) are the keys to optimally control the internal computation flow which can be formulated by Eq.\ref{Eq:lstm}
\begin{equation}
    \begin{aligned}
         i_{t} &= \sigma(x_{t}\cdot{\textbf{W}_{xi}}+h_{t-1}\cdot{\textbf{W}_{hi}}+b_{i}) \\
         f_{t} &= \sigma(x_{t}\cdot{\textbf{W}_{xf}}+h_{t-1}\cdot{\textbf{W}_{hf}}+b_{f}) \\ 
         o_{t} &= \sigma(x_{t}\cdot{\textbf{W}_{xo}}+h_{t-1}\cdot{\textbf{W}_{ho}}+b_{o}) \\
         g_{t} &= \textit{tanh}(x_{t}\cdot{\textbf{W}_{xg}}+h_{t-1}\cdot{\textbf{W}_{hg}}+b_{g})  \\
         c_{t} &= f_{t} \otimes c_{t-1} + i_{t} \otimes g_{t} \\
         h_{t} &=o_{t} \otimes(c_{t})
 \label{Eq:lstm}
    \end{aligned}
\end{equation}

where $x_{t}, h_{t}$ refer to the input, hidden state at step $t$;  $x_{t-1}, h_{t-1}$ refer to the input, hidden state at step $t-1$;  $\otimes$ is element-wise multiplication and $\cdot$ is matrix multiplication; $\sigma(\cdot)$ is sigmoid function and $\textit{tanh}(\cdot)$ is hyperbolic tangent function; $\textbf{W}$ and $b$ refer to parameters within the gates to optimize how much of information should be let through. 

Despite the outstanding performance of deeply stacked LSTMs, the subsequent cost that comes with it is unacceptable. Decreasing the number of parameters inside cells is a promising way to fulfill much deeper LSTMs with as many parameters as one layer of LSTM. Essentially, the learning process of those four gates can be treated as four fully-connected layers which are prone to be over-parameterized. Especially, in order to remember the information for a long period of time, plenty of cells needed to be connected sequentially and thus, the reuse of these gates leads to unnecessary computation cost.

To apply SET to these four gates, we first use an Erd{\H{o}}s-R{\'e}nyi topology to randomly create sparse layers which  replace the FC layers corresponding to the four gates. Then, we apply the rewiring process to dynamically prune and add connections to optimize the computation flow. After learning, different gates are able to learn their specific sparse structure according to their roles. We illustrate the SET-LSTM diagram in Figure \ref{Fig:SparseLstmCell}.

\subsection{SET-LSTM Embedding}
Word embedding has been widely applied in natural language processing tasks to improve the performance of the models with discrete inputs such as words, as one of the distributed word representations. Recently, neural network architectures have attracted tremendous attention at word embedding, among them, CBOW and the skip-gram model in word2vec\cite{mikolov2013distributed} are the most well-known, as they can not only project the words in a vector space but can preserve the syntactic and semantic relations between the words.

The conventional word embedding methods project words to dense vectors, as shown in Figure \ref{Fig:dense_embedding}. The word embedding is obtained by the product of the input, a ``one-hot'' encoded vector (a zeros vector in which only one position is 1), with an embedding matrix $\textbf{W}_{E} \in \mathbb{R}^{V \times{D}}$, where  $D$ is the dimension of the word embedding and $V$ is the total number of words. Practically, this embedding layer is the largest layer in most LSTM models with a huge number of parameters ($DV$). Thus, it is desirable to apply SET to the embedding layer. 

Same as in the implementation of SET-LSTM cells, we replace the dense rows of matrix $\textbf{W}_{E}$ with sparse ones and during training, we apply the weight-removal and weight-addition steps to adjust the topology. We illustrate our SET-LSTM embedding layer in Figure \ref{Fig:sparse_embedding}. 

\section{Experimental Results}
We evaluate our method on four sentiment analysis datasets: IMDB \cite{maas2011learning}, Sanders Corpus Twitter\footnote{ http://www.sananalytics.com/lab/twitter-sentiment/}, Yelp 2018\footnote{https://www.yelp.com/dataset/challenge} and Amazon Fine Food Reviews\footnote{https://snap.stanford.edu/data/web-FineFoods.html}.

\subsection{Experimental Setup}
We randomly choose 80\% of the data as training set and the remaining 20\% as testing set for all datasets, except IMDB (25000 for training and 25000 for testing). For the sake of convenience, on all datasets, we set the sparsity hyperparameter to be $\epsilon = 10$, which means there are $10\times{(n^{k}+n^{k-1})}$ connections between layer $k$ and layer $k-1$; we set the dimension of word embedding to be 256, the hidden state of LSTM unit to be 256; and the number of words in each sentence is 100 and the total number of words in embedding is 20000. The rewire rate $\zeta = 0.2$ for Yelp 2018 and Amazon, $\zeta = 0.6$ for Twitter and $\zeta = 0.4$ for IMDB; Additionally, the mini-batch size is 64 for Twitter, Yelp and Amazon, and 256 for IMDB. We train the models using Adam optimizer with the learning rate of 0.001 for Twitter and Amazon, 0.01 for Yelp 2018, and 0.0005 for IMDB. 

We compare SET-LSTM with fully-connected LSTM, and SETC-LSTM (SET-LSTM with sparse LSTM cells and a FC embedding layer). In order to make a fair comparison, all these three models have the same hyperparameters and are implemented with the same architecture, that is, one embedding layer, one LSTM layer followed by one dense output layer. We didn't make the output layer sparse since its amount is negligible in comparison with the total number of parameters. We didn't compare our method with the other recent directly sparse methods such as Nest ,DEEP R, and dynamic sparse reparameterization. Essentially, Nest actually does not limit the number of parameters to a strict budget, as it grows a small network to a large one and then prunes it down. The comparison between DEEP R and SET has been made in \cite{mostafa2019parameter} and it shows for WRN-28-2 on CIFAR10 that SET is able to achieve better performance than DEEP R with four times lower computational overhead of rewiring process during training. In terms of dynamic reparameterization, its differences from SET are only the thresholds to remove weights and the way to reallocate the connections across layers. 

\subsection{Results}

\begin{table*}[t]
\caption{Sentiment analysis test accuracy and sparsity on IMDB, Twitter, Yelp 2018 and Amazon}
\label{tabel_accuracy}
\vskip 0.15in
\begin{center}
\begin{tabular}{lrrrrrr}
\toprule
Methods & IMDB (\%) & Twitter (\%) & Yelp 2018 (\%) & Amazon (\%) & Parameters (\#) & Sparsity (\%)\\
\midrule
LSTM & 85.26 & 77.79 & 63.36 & 81.88 & 5,645,312 & 0\\
SETC-LSTM & 85.42($\pm0.10$) & 77.59($\pm0.53$) & 67.82($\pm0.33$) & 81.52($\pm0.12$) & 5,161,012 & 8.58\\
SET-LSTM & 86.04($\pm0.22$) & 79.22($\pm0.56$) & 68.00($\pm0.18$) & 80.52($\pm0.15$) & 243,442 & 95.69  \\
\hline \\
\end{tabular}
\end{center}
\vskip -0.1in
\end{table*}

The experimental results are reported in Table \ref{tabel_accuracy}. Every accuracy is collected and averaged from five different trials, as the topology and weights are initialized randomly. The table shows that only by applying SET to LSTM cells, SETC-LSTM is able to increase the accuracy of fully connected LSTM by 0.16\% and 4.46\% on IMDB and Yelp 2018, respectively, whereas it causes negligible decreases on the other two datasets (0.20\% for Twitter and 0.36\% for Amazon). However, further taking both the LSTM cells and embedding layer into account, SET-LSTM can outperform LSTM on three datasets, by 0.78\% for IMDB, by 1.43\% for Twitter and by 4.64\% for Yelp 2018, respectively. The only dataset that SET-LSTM does not increase the accuracy is Amazon with 1.36\% loss of accuracy. We mention here that the accuracy on Amazon can be improved by searching for the best hyperparameters, but it was out of the goal of this paper.

Given the large number of parameters of the embedding layer, the sparsity caused by LSTM cells is very limited (8.58\%). However, after we apply SET to the embedding layer, the sparsity increases dramatically and reaches 95.69\%. We didn't sparsify the connections of the output layer, because the number is too small to influence the overall sparsity level. Since for all datasets, the architecture and the hyperparameters that determine the level of sparsity such as $\epsilon$, the number of embedding features, the number of hidden units and the word number of embedding are the same, the sparsity level is the same. 

Beside this, we are also interested if SET-LSTM is still trainable under extreme sparsity. To do this, we set the sparsity to an extreme level (99.1\%) and we compare our algorithm with fully-connected LSTM. Due to time constraints, we only test our approach on IMDB, Twitter and Yelp 2018. The results are shown in Table \ref{tabel_extremesparsity}. With more than 99\% sparsity, our method is still able to find a good sparse topology with competitive performance.
\begin{table}[t]
\caption{Sentiment analysis test accuracy of SET-LSTM under extreme sparsity (99.1\%) on IMDB, Twitter, Yelp 2018}
\label{tabel_extremesparsity}
\vskip 0.15in
\begin{center}
\begin{tabular}{lrrr}
\toprule
Methods & IMDB(\%) & Twitter(\%) & Yelp 2018(\%) \\
\midrule
LSTM & 85.26 & 77.79 & 63.36\\
\hline
SET-LSTM & 85.05 & 78.85 & 67.82\\
\hline 
\end{tabular}
\end{center}
\vskip -0.1in
\end{table}

\begin{figure*}[ht]
\vskip 0in
\begin{center}
\centerline{\includegraphics[width=150mm,height=100mm]{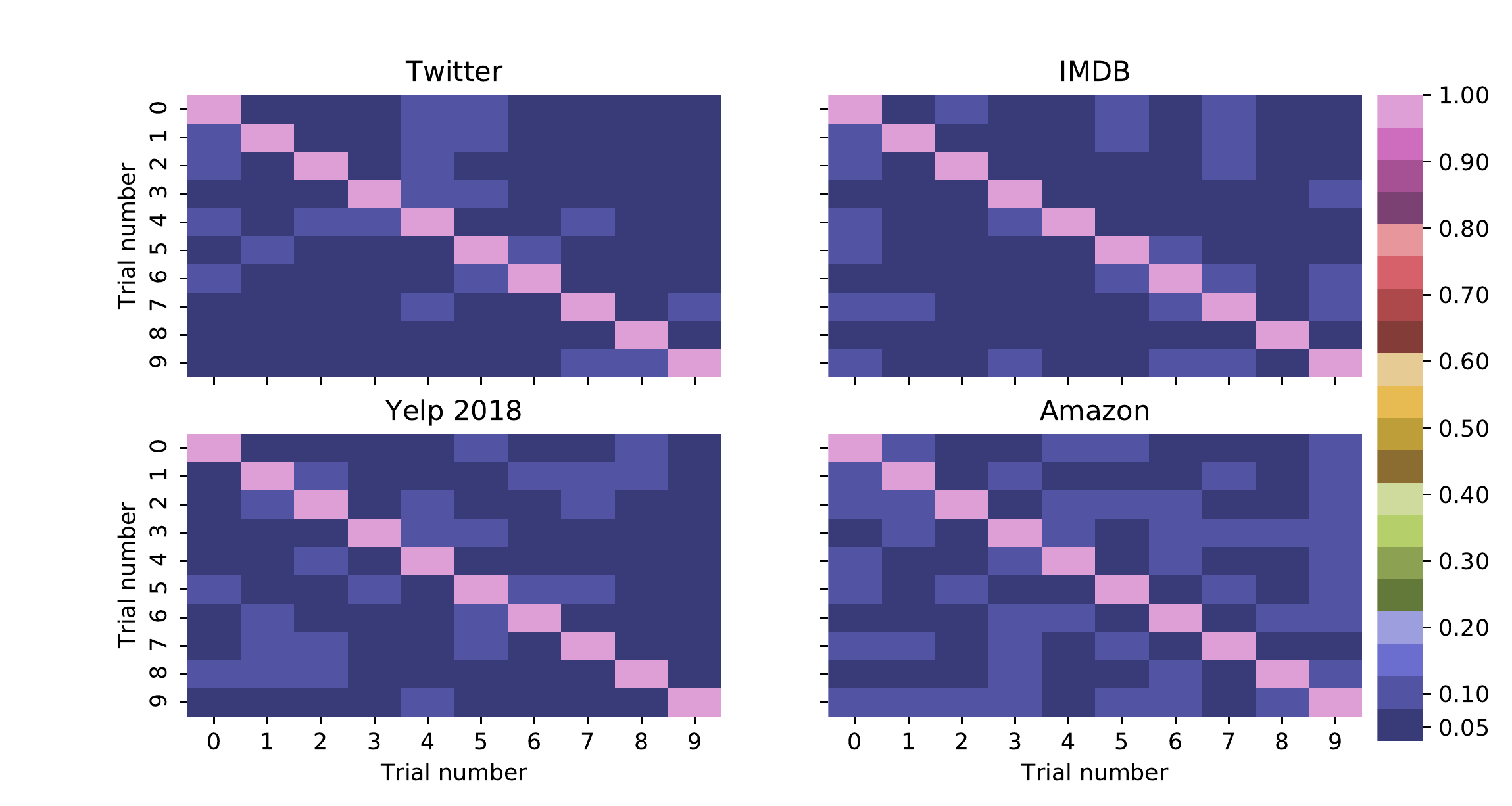}}
\caption{Similarity matrices of LSTM cells for Twitter, IMDB, Yelp 2018 and Amazon}
\label{Fig:similarity_lstm}
\end{center}
\vskip -0.2in
\end{figure*}

\subsection{Analysis}
\begin{figure*}[ht]
\vskip 0in
\begin{center}
\centerline{\includegraphics[width=150mm,height=100mm]{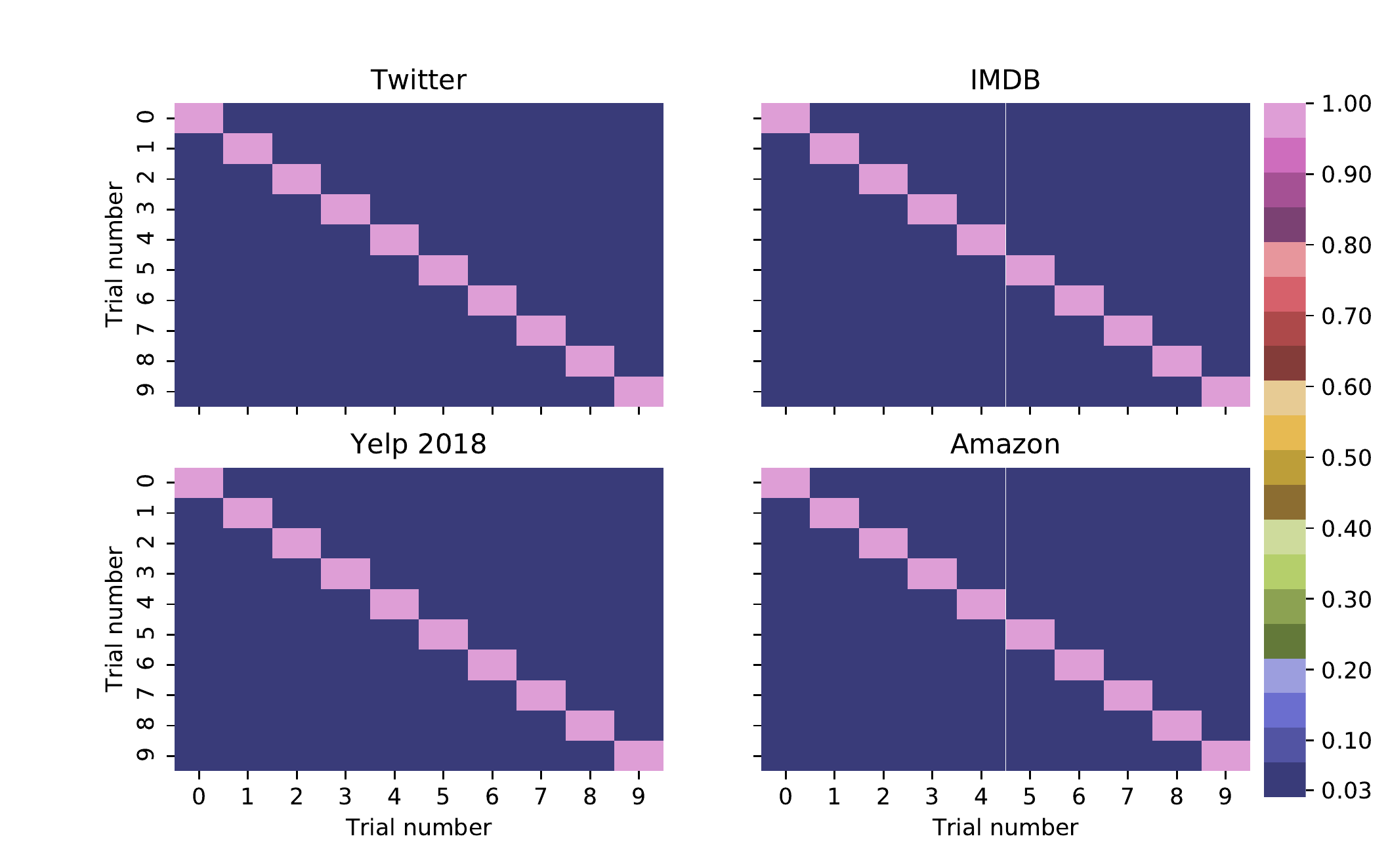}}
\caption{Similarity matrices of LSTM embedding layer for Twitter, IMDB, Yelp 2018 and Amazon}
\label{Fig:similarity_embedding}
\end{center}
\vskip -0.5in
\end{figure*}

\begin{figure}[ht]
\vskip 0.2in
\begin{center}
\centerline{\includegraphics[width=\columnwidth]{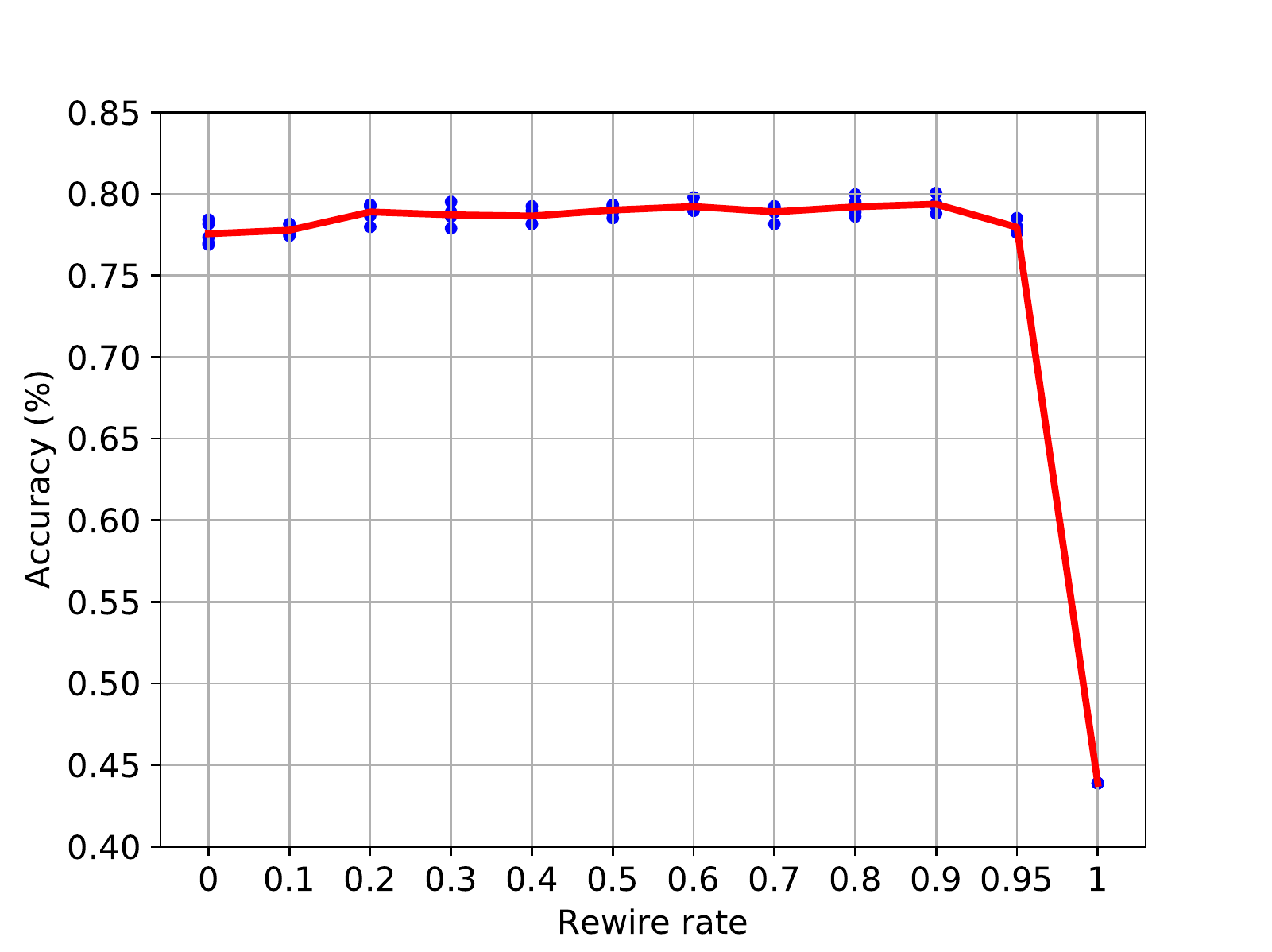}}
\caption{Test accuracy with different rewire rates $\zeta$  on Twitter.}
\label{Fig:rewire_twitter}
\end{center}
\vskip -0.3in
\end{figure}

\begin{figure}[ht]
\vskip 0.2in
\begin{center}
\centerline{\includegraphics[width=\columnwidth]{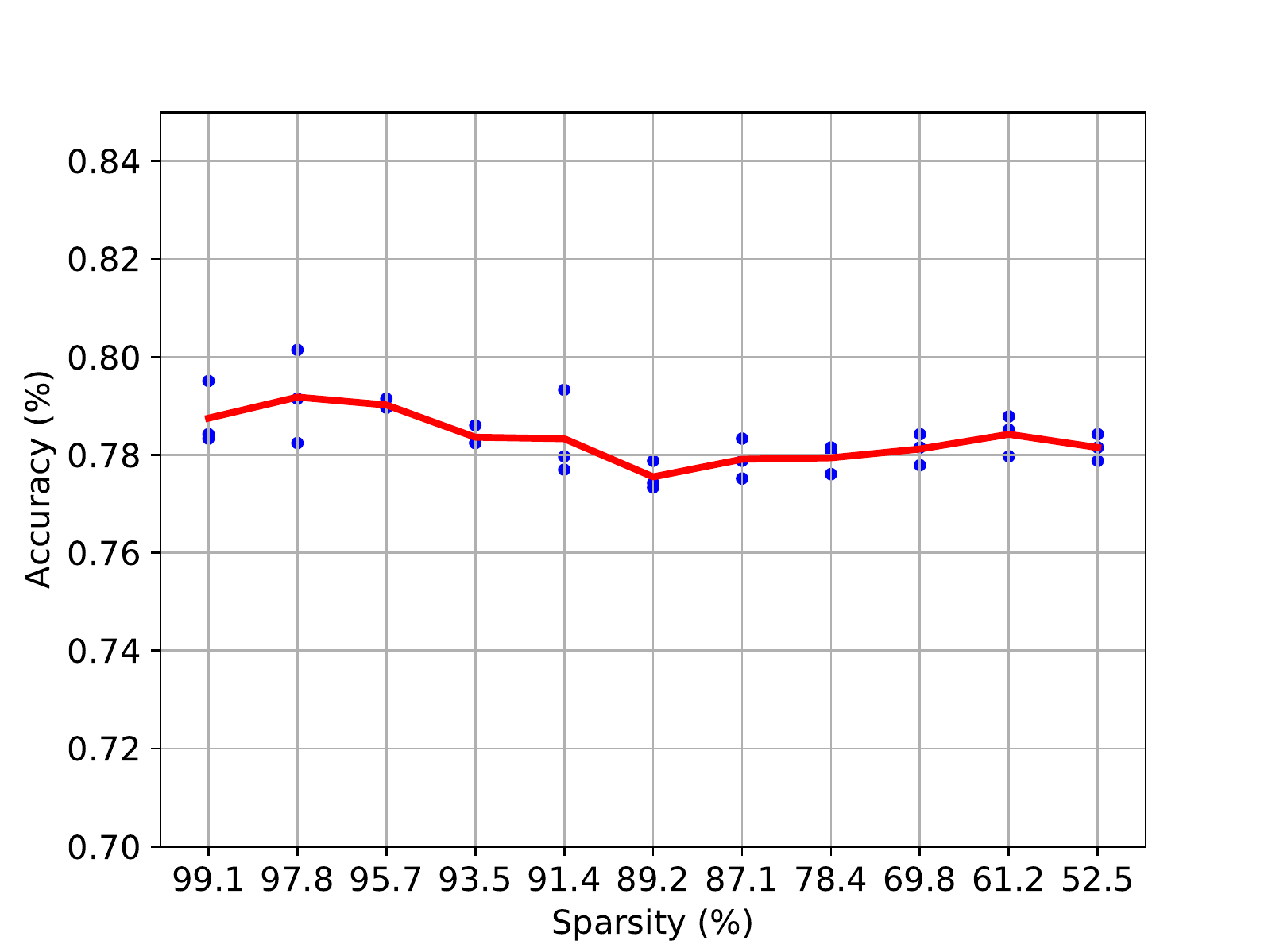}}
\caption{Test accuracy with different sparsity levels on Twitter}
\label{Fig:sparsity}
\end{center}
\vskip -0.2in
\end{figure}
It has been shown that SET is capable of reducing the size of network quadratically with no decrease in accuracy \cite{mocanu2018scalable, mostafa2019parameter}, whereas there is no convincing theoretical explanation which can uncover the secret of this phenomenon. Here, we give a plausible rationale, that is, there are plenty of different sparse topologies across layers (local optima) that can properly represent one fully connected overparameterized neural network. This means that starting from different sparse topologies, different trials (different runs of SET-LSTM) will evolve toward different topologies, and all those topologies can be good local optima. To support this hypothesis, we do 10 trials on each dataset, and we calculate the similarity of their best topologies (corresponding to their best accuracy). The similarity of topology $a$ with regard to $b$ is defined as:
\begin{equation}
 S_{ab}=\frac{n_{ab}}{n_{a}}
 \label{Eq:similarity}
\end{equation}
where $n_{ab}$ is the number of common connections in both topologies, i.e. $a$ and $b$; and $n_a$ is the total number of connections in topology $a$. We treat the connection (${W}^k_{ij}$) as a common connection when both topologies contain a connection between the $i^{th}$ neuron of the layer k-1 and the $j^{th}$ neuron of the layer k. The similarity of LSTM cells and the embedding layer are shown in Figure \ref{Fig:similarity_lstm} and Figure \ref{Fig:similarity_embedding}, respectively. It can be observed that for Twitter the similarity of the different topologies is very small, around 8\% for LSTM cells and 4.5\% for the embedding layer. This finding is consistent across other datasets. The evidence supports the rationale that sparse neural networks provide many low-dimensional structures to substitute the optima of the overparameterized deep neural networks which usually are high-dimensional manifolds. This hypothesis is also consistent with the point of view of \cite{cooper2018loss}, which shows that the locus of a global minima of an overparameterized neural network is a high-dimensional subset of $\mathbb{R}^n$.

\begin{table*}[t]
\caption{The test accuracy of ten trials for IMDB, Twitter, Yelp 2018 and Amazon, in percentage.}
\label{table_trials}
\vskip 0.15in
\begin{center}
\begin{tabular}{lcccccccccc}
\toprule
 & Trail1 & Trail2 & Trail3 & Trail4 & Trail5 & Trail6 & Trail7 & Trail8 & Trail9 & Trial10\\
\midrule
IMDB & 85.77 & 86.01 & 86.00 & 86.16 & 85.97 & 85.96 & 85.90 & 86.03 & 85.80 & 86.00 \\
Twitter & 78.97 & 79.15 & 78.97 & 79.78 & 79.24 & 78.24 & 80.14 & 79.14 & 79.24 & 79.33 \\
Yelp 2018 & 68.12 & 67.89 & 68.12 & 67.84 & 68.02 & 68.25 & 68.00 & 68.13 & 68.00 & 67.94 \\
Amazon & 80.20 & 80.56 & 79.69 & 80.78 & 80.28 & 79.85 & 80.78 & 79.95 & 80.12 & 80.52 \\
\hline \\
\end{tabular}
\end{center}
\vskip -0.1in
\end{table*}
\section{Extra Analysis with Twitter}
In this section, we do several extra experiments on the Sanders Corpus Twitter dataset to gain more insights into the details of SET-LSTM. 
This data set consists of 5513 tweets manually labeled with regard to one of four topics (Apple, Google, Microsoft and Twitter). Out of 5513 tweets, there are 654 negative, 2,503 neutral, 570 positive and 1,786 irrelevant tweets. 

\textbf{Rewire rate} As a hyperparameter of SET-LSTM, the rewire rate determines how many connections should be removed after each epoch. We examine 11 different rewire rates $\zeta$, 5 trials for each $\zeta$, to find the best rewire rate for Twitter. The comparison is reported in Figure \ref{Fig:rewire_twitter} showing that the rewire rate has a relatively wide range of safe options. The best choice of $\zeta$ is 0.9 whose average accuracy is 79.37\%. It seems that by keeping just 10 percent of the connections in each epoch, it is enough to fit the Twitter dataset. 

\textbf{The importance of initialization} Considering that our evolutionary training dynamically forces the topology to a local optimal one, it is interesting to check whether using a fixed optimal topology learned by SET-LSTM will reach the same accuracy or not. We use two methods to examine this. One uses a fixed optimal topology learned by a previous trial (whose accuracy is 78.89\%), and with randomly initialized weights values. The other one also uses the same topology but the weights values are initialized with the ones of the original trial. The results of this experiment are shown in Table \ref{table_initialization}. When randomly initialized, the network with a fixed topology is not able to achieve the same accuracy, whereas using the same initialization it can even achieve better accuracy. This suggests that the optimization all-together of weights and topology done by the evolutionary process during training is a critical process in finding optimal sparse topologies, while a good initialization is very important for sparse networks. The latter aspect also matches the findings from \cite{frankle2018lottery} which state that the initialization of a winning ticket (sparse topology) is important to its success, while the evolutionary process from SET-LSTMs ensures a way to \textit{always} find the winning ticket.

\begin{table}[t]
\caption{The performance of the SET-LSTM for Twitter when the topology is fixed with an optimal one, in percentage.}
\label{table_initialization}
\vskip 0.15in
\begin{center}
\begin{small}
\begin{tabular}{lccc}
\toprule
 & SET-LSTM & Randomly  & Same  \\
 &       & initialization & initialization \\
\midrule
Twitter & 78.89 & 77.97($\pm1.00$) & 78.91($\pm0.40$) \\
\hline \\
\end{tabular}
\end{small}
\end{center}
\vskip -0.4in
\end{table}

\textbf{The trade-off between sparsity and performance}
Basically, there is a trade-off between the sparsity level and classification performance for sparse neural networks. If the network is too sparse, it will not have sufficient capacity to fit the dataset, but if the network is too dense, the decrease in the number of parameters will be too small to influence the computation and memory requests. In order to find the safe choice of sparsity, we run an experiment three times for 7 different $\epsilon$. The results are reported in Figure \ref{Fig:sparsity}. 
It is worth noting that,for extreme sparsity, when $\epsilon = 2$ (sparsity is equal to 99.1\%), the accuracy (78.75\%) is still higher than LSTM (77.19\%). Moreover, it is interesting to see that when the sparsity level goes down under 90\% the accuracy is also going down, this being in line with our observation that usually sparse networks with adaptive sparse connectivity perform better than fully connected networks.

\section{Conclusions}
In this paper, we propose SET-LSTM to deal with the situation where the budget of parameters is strictly limited. By applying SET to the LSTM cells and the embedding layer, we are not only able to eliminate more than 99\% parameters, but to achieve better performance on three datasets. Additionally, we find that the optimal topology learned by SET are very different from each other. The potential explanation is that SET-LSTM can find many amenable low-dimensional sparse topologies, being capable of replacing efficiently the costly optimization of overparameterized dense neural networks.

Up to now, we only evaluate our proposed method on sentiment analysis text datasets. In future work, we intend to understand deeper why SET-LSTM is able to reach better performance than its fully connected counterparts. Also, we intend to implement a vanilla SET-LSTM using just sparse data structures to take advantage of its full potential. On the application side, we intend to use SET-LSTM for other types of time series problems, e.g. speech recognition.




\bibliography{example_paper}

\begin{thebibliography}{23}
\providecommand{\natexlab}[1]{#1}
\providecommand{\url}[1]{\texttt{#1}}
\expandafter\ifx\csname urlstyle\endcsname\relax
  \providecommand{\doi}[1]{doi: #1}\else
  \providecommand{\doi}{doi: \begingroup \urlstyle{rm}\Url}\fi

\bibitem[Bellec et~al.(2017)Bellec, Kappel, Maass, and
  Legenstein]{bellec2017deep}
Bellec, G., Kappel, D., Maass, W., and Legenstein, R.
\newblock Deep rewiring: Training very sparse deep networks.
\newblock \emph{arXiv preprint arXiv:1711.05136}, 2017.

\bibitem[Cooper(2018)]{cooper2018loss}
Cooper, Y.
\newblock The loss landscape of overparameterized neural networks.
\newblock \emph{arXiv preprint arXiv:1804.10200}, 2018.

\bibitem[Dai et~al.(2017)Dai, Yin, and Jha]{dai2017nest}
Dai, X., Yin, H., and Jha, N.~K.
\newblock Nest: a neural network synthesis tool based on a grow-and-prune
  paradigm.
\newblock \emph{arXiv preprint arXiv:1711.02017}, 2017.

\bibitem[Frankle \& Carbin(2018)Frankle and Carbin]{frankle2018lottery}
Frankle, J. and Carbin, M.
\newblock The lottery ticket hypothesis: Finding sparse, trainable neural
  networks.
\newblock 2018.
\newblock URL \url{https://openreview.net/forum?id=rJl-b3RcF7}.

\bibitem[Giles \& Omlin(1994)Giles and Omlin]{giles1994pruning}
Giles, C.~L. and Omlin, C.~W.
\newblock Pruning recurrent neural networks for improved generalization
  performance.
\newblock \emph{IEEE transactions on neural networks}, 5\penalty0 (5):\penalty0
  848--851, 1994.

\bibitem[Graves et~al.(2013)Graves, Jaitly, and Mohamed]{graves2013hybrid}
Graves, A., Jaitly, N., and Mohamed, A.-r.
\newblock Hybrid speech recognition with deep bidirectional lstm.
\newblock In \emph{Automatic Speech Recognition and Understanding (ASRU), 2013
  IEEE Workshop on}, pp.\  273--278. IEEE, 2013.

\bibitem[Han et~al.(2015)Han, Pool, Tran, and Dally]{han2015learning}
Han, S., Pool, J., Tran, J., and Dally, W.
\newblock Learning both weights and connections for efficient neural network.
\newblock In \emph{Advances in neural information processing systems}, pp.\
  1135--1143, 2015.

\bibitem[Han et~al.(2017)Han, Kang, Mao, Hu, Li, Li, Xie, Luo, Yao, Wang,
  et~al.]{han2017ese}
Han, S., Kang, J., Mao, H., Hu, Y., Li, X., Li, Y., Xie, D., Luo, H., Yao, S.,
  Wang, Y., et~al.
\newblock Ese: Efficient speech recognition engine with sparse lstm on fpga.
\newblock In \emph{Proceedings of the 2017 ACM/SIGDA International Symposium on
  Field-Programmable Gate Arrays}, pp.\  75--84. ACM, 2017.

\bibitem[Jouppi et~al.(2017)Jouppi, Young, Patil, Patterson, Agrawal, Bajwa,
  Bates, Bhatia, Boden, Borchers, et~al.]{jouppi2017datacenter}
Jouppi, N.~P., Young, C., Patil, N., Patterson, D., Agrawal, G., Bajwa, R.,
  Bates, S., Bhatia, S., Boden, N., Borchers, A., et~al.
\newblock In-datacenter performance analysis of a tensor processing unit.
\newblock In \emph{Computer Architecture (ISCA), 2017 ACM/IEEE 44th Annual
  International Symposium on}, pp.\  1--12. IEEE, 2017.

\bibitem[Kuchaiev \& Ginsburg(2017)Kuchaiev and
  Ginsburg]{kuchaiev2017factorization}
Kuchaiev, O. and Ginsburg, B.
\newblock Factorization tricks for lstm networks.
\newblock \emph{arXiv preprint arXiv:1703.10722}, 2017.

\bibitem[Lee et~al.(2018)Lee, Ajanthan, and Torr]{lee2018snip}
Lee, N., Ajanthan, T., and Torr, P.~H.
\newblock Snip: Single-shot network pruning based on connection sensitivity.
\newblock \emph{arXiv preprint arXiv:1810.02340}, 2018.

\bibitem[Lobacheva et~al.(2017)Lobacheva, Chirkova, and
  Vetrov]{lobacheva2017bayesian}
Lobacheva, E., Chirkova, N., and Vetrov, D.
\newblock Bayesian sparsification of recurrent neural networks.
\newblock \emph{arXiv preprint arXiv:1708.00077}, 2017.

\bibitem[Lu et~al.(2016)Lu, Sindhwani, and Sainath]{lu2016learning}
Lu, Z., Sindhwani, V., and Sainath, T.~N.
\newblock Learning compact recurrent neural networks.
\newblock \emph{arXiv preprint arXiv:1604.02594}, 2016.

\bibitem[Maas et~al.(2011)Maas, Daly, Pham, Huang, Ng, and
  Potts]{maas2011learning}
Maas, A.~L., Daly, R.~E., Pham, P.~T., Huang, D., Ng, A.~Y., and Potts, C.
\newblock Learning word vectors for sentiment analysis.
\newblock In \emph{Proceedings of the 49th annual meeting of the association
  for computational linguistics: Human language technologies-volume 1}, pp.\
  142--150. Association for Computational Linguistics, 2011.

\bibitem[Mikolov et~al.(2013)Mikolov, Sutskever, Chen, Corrado, and
  Dean]{mikolov2013distributed}
Mikolov, T., Sutskever, I., Chen, K., Corrado, G.~S., and Dean, J.
\newblock Distributed representations of words and phrases and their
  compositionality.
\newblock In \emph{Advances in neural information processing systems}, pp.\
  3111--3119, 2013.

\bibitem[Mocanu et~al.(2018)Mocanu, Mocanu, Stone, Nguyen, Gibescu, and
  Liotta]{mocanu2018scalable}
Mocanu, D.~C., Mocanu, E., Stone, P., Nguyen, P.~H., Gibescu, M., and Liotta,
  A.
\newblock Scalable training of artificial neural networks with adaptive sparse
  connectivity inspired by network science.
\newblock \emph{Nature Communications}, 9\penalty0 (1):\penalty0 2383, 2018.

\bibitem[Mostafa \& Wang(2019)Mostafa and Wang]{mostafa2019parameter}
Mostafa, H. and Wang, X.
\newblock Parameter efficient training of deep convolutional neural networks by
  dynamic sparse reparameterization, 2019.
\newblock URL \url{https://openreview.net/forum?id=S1xBioR5KX}.

\bibitem[Narang et~al.(2017)Narang, Elsen, Diamos, and
  Sengupta]{narang2017exploring}
Narang, S., Elsen, E., Diamos, G., and Sengupta, S.
\newblock Exploring sparsity in recurrent neural networks.
\newblock \emph{arXiv preprint arXiv:1704.05119}, 2017.

\bibitem[Sutskever et~al.(2014)Sutskever, Vinyals, and
  Le]{sutskever2014sequence}
Sutskever, I., Vinyals, O., and Le, Q.~V.
\newblock Sequence to sequence learning with neural networks.
\newblock In \emph{Advances in neural information processing systems}, pp.\
  3104--3112, 2014.

\bibitem[Tian et~al.(2017)Tian, Zhang, Ma, He, Wei, Wu, Situ, Li, and
  Zhang]{tian2017deep}
Tian, X., Zhang, J., Ma, Z., He, Y., Wei, J., Wu, P., Situ, W., Li, S., and
  Zhang, Y.
\newblock Deep lstm for large vocabulary continuous speech recognition.
\newblock \emph{arXiv preprint arXiv:1703.07090}, 2017.

\bibitem[Wen et~al.(2017)Wen, He, Rajbhandari, Zhang, Wang, Liu, Hu, Chen, and
  Li]{wen2017learning}
Wen, W., He, Y., Rajbhandari, S., Zhang, M., Wang, W., Liu, F., Hu, B., Chen,
  Y., and Li, H.
\newblock Learning intrinsic sparse structures within long short-term memory.
\newblock \emph{arXiv preprint arXiv:1709.05027}, 2017.

\bibitem[Yang et~al.(2016)Yang, Yang, Dyer, He, Smola, and
  Hovy]{yang2016hierarchical}
Yang, Z., Yang, D., Dyer, C., He, X., Smola, A., and Hovy, E.
\newblock Hierarchical attention networks for document classification.
\newblock In \emph{Proceedings of the 2016 Conference of the North American
  Chapter of the Association for Computational Linguistics: Human Language
  Technologies}, pp.\  1480--1489, 2016.

\bibitem[Zen et~al.(2016)Zen, Agiomyrgiannakis, Egberts, Henderson, and
  Szczepaniak]{zen2016fast}
Zen, H., Agiomyrgiannakis, Y., Egberts, N., Henderson, F., and Szczepaniak, P.
\newblock Fast, compact, and high quality lstm-rnn based statistical parametric
  speech synthesizers for mobile devices.
\newblock \emph{arXiv preprint arXiv:1606.06061}, 2016.

\end{thebibliography}
\bibliographystyle{icml2019}

\end{document}